\documentclass{article}
\usepackage{amsmath}

\usepackage{PRIMEarxiv}

\usepackage[utf8]{inputenc} 
\usepackage[T1]{fontenc}    
\usepackage{hyperref}       
\usepackage{url}            
\usepackage{booktabs}       
\usepackage{amsfonts}       
\usepackage{nicefrac}       
\usepackage{microtype}      
\usepackage{lipsum}
\usepackage{fancyhdr}       
\usepackage{graphicx}       
\graphicspath{{media/}}     

\title{LogitLens4LLMs: Extending Logit Lens Analysis to Modern Large Language Models
}

\author{
 Zhenyu Wang \\
  Independent Researcher \\
  Beijing, China\\
  \texttt{zhenyu\_wang02@outlook.com} \\
}

\begin{document}
\maketitle

\begin{abstract}
This paper introduces LogitLens4LLMs, a toolkit that extends the Logit Lens technique to modern large language models. While Logit Lens has been a crucial method for understanding internal representations of language models, it was previously limited to earlier model architectures. Our work overcomes the limitations of existing implementations, enabling the technique to be applied to state-of-the-art architectures (such as Qwen-2.5 and Llama-3.1) while automating key analytical workflows. By developing component-specific hooks to capture both attention mechanisms and MLP outputs, our implementation achieves full compatibility with the HuggingFace transformer library while maintaining low inference overhead. The toolkit provides both interactive exploration and batch processing capabilities, supporting large-scale layer-wise analyses. Through open-sourcing our implementation, we aim to facilitate deeper investigations into the internal mechanisms of large-scale language models. The toolkit is openly available at \url{https://github.com/zhenyu-02/LogitLens4LLMs}.
\end{abstract}

\section{Introduction}
The Logit Lens technique\cite{nostalgebraist2020interpreting}, first proposed in 2020, has emerged as a crucial tool for understanding the internal representations of large language models (LLMs). By projecting intermediate layer activations through the model's final language modeling head, researchers can observe how token prediction distributions evolve throughout the network's processing hierarchy. This method has proven invaluable for analyzing reasoning pathways \cite{zhaoyi2024understanding}, multilingual representations \cite{wendler2024llamasworkenglishlatent}, and safety-critical behaviors \cite{ashkan2024mechanistic}.\\

Despite its widespread adoption, current implementations remain limited to older architectures like GPT-2 and early Llama variants\cite{panickssery2023decoding}. Contemporary models exhibit substantially different scaling properties and architectural modifications that render existing tools inadequate. Moreover, the manual processing required for data collection and analysis creates significant barriers to systematic investigation. Our work addresses these limitations through LogitLens4LLMs, which extends the methodology to state-of-the-art models while automating key analytical workflows.

\section{Related Work}
\label{sec:related-work}
The original Logit Lens concept demonstrated that meaningful token predictions could be extracted from intermediate transformer layers \cite{nostalgebraist2020interpreting}. Subsequent improvements introduced the Tuned Lens \cite{belrose2023elicitinglatentpredictionstransformers}, which applies learned transformations to intermediate activations for clearer interpretation. Parallel work developed complementary techniques like Patchscopes \cite{ghandeharioun2024patchscopesunifyingframeworkinspecting} and gradient projection methods \cite{katz2024backwardlensprojectinglanguage}, creating a rich ecosystem of transformer analysis tools.\\
\\
Recent applications have revealed fundamental insights about LLM internals. Wu\cite{wu2024semantichubhypothesislanguage} used Logit Lens to identify cross-lingual semantic hubs, while Jia\cite{jaturong2025an} employed it to study prediction refinement dynamics. Safety researchers \cite{ashkan2024mechanistic} and alignment investigators \cite{bhalla2025unifyinginterpretabilitycontrolevaluation} have leveraged these techniques to localize critical model behaviors. However, all such studies remain constrained by the limited model support of existing implementations.

\section{System Design}
The Logit Lens methodology operates on the principle that intermediate layer activations contain progressively refined predictions about subsequent tokens. Formally, given a transformer model with $L$ layers, the predictive distribution at layer $l$ can be expressed as:

\begin{equation}
p_l(x_{t+1}|x_{\leq t}) = \text{softmax}(W_{\text{head}} \cdot \text{Norm}(h_l^{(t)}))
\end{equation}

where $h_l^{(t)}$ denotes the hidden state at layer $l$ for position $t$, $\text{Norm}$ represents the model's final layer normalization, and $W_{\text{head}}$ is the language modeling head's weight matrix.

Our implementation extends this core principle to Qwen-2.5 and Llama-3.1 through three architectural adaptations. We develop component-specific hooks to capture attention mechanisms and MLP outputs simultaneously:

\begin{equation}
h^{(l+1)} = f_{\text{attn}}(h^{(l)}) + f_{\text{mlp}}(\text{Norm}(h^{(l)} + f_{\text{attn}}(h^{(l)})))
\end{equation}

where $f_{\text{attn}}$ and $f_{\text{mlp}}$ represent the attention and MLP sublayers respectively. Our wrapper architecture preserves the original computation graph while intercepting intermediate results at four critical points: post-attention, intermediate residual, MLP output, and block output.

\begin{figure}[ht]
\centering
\includegraphics[width=1.2\textwidth, height=1.2\textwidth, keepaspectratio]{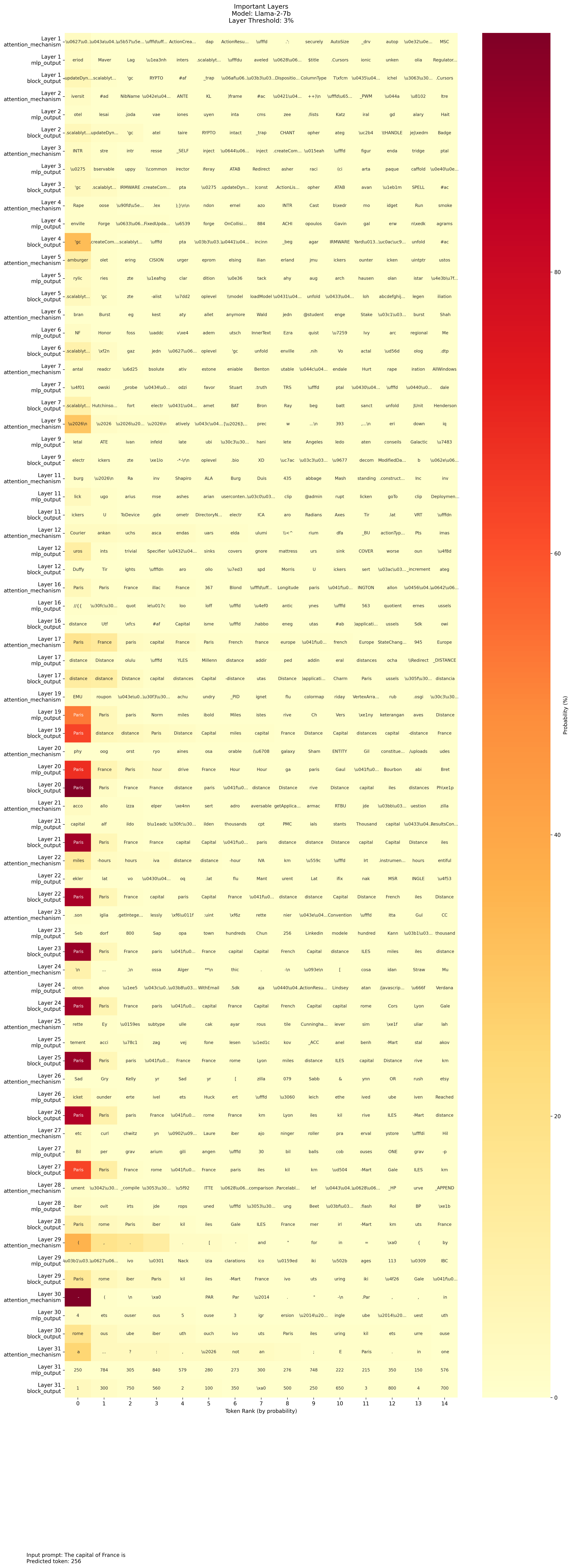}
\caption{Prediction evolution across layers in Llama-2-7B for the prompt "The capital of France is "}
\label{fig:heatmap}
\end{figure}

The visualization system generates layer-wise prediction heatmaps through an automated pipeline:

\begin{equation}
H_{i,j} = \frac{1}{Z}\sum_{k=1}^K \mathbb{I}(\text{token}_k \in T_j) \cdot p_l(\text{token}_k|x_{\leq t})
\end{equation}

where $H_{i,j}$ represents the heatmap value at layer $i$ for token $j$, $T_j$ denotes the set of top-$K$ predicted tokens, and $Z$ is a normalization constant. Figure~\ref{fig:heatmap} demonstrates this visualization for a multi-step prediction task, showing how early layers (5-15) establish semantic context while later layers (20-30) resolve lexical specificity.

Our implementation achieves full compatibility with HuggingFace's transformer library through gradient-preserving hooks that maintain small overhead during inference. The modular architecture supports both interactive exploration via Jupyter notebooks and batch processing for large-scale layer-wise analyses.

\section{Conclusion}
LogitLens4LLMs significantly advances the practical utility of activation decoding techniques for modern language models. By extending support to cutting-edge architectures and automating previously manual processes, our toolkit enables large-scale comparative studies of model internals. The included visualization capabilities further lower the barrier to mechanistic analysis, particularly for researchers investigating emergent capabilities in multi-billion parameter systems. We open-source our implementation at \url{https://github.com/zhenyu-02/LogitLens4LLMs} to support ongoing interpretability research.

\section{Acknowledgement}
The author would like to express gratitude to Zaoyi Li and Herman (Xiaohongshu user) for their valuable discussions on extending Logit Lens applications to various models. Special thanks to Nina Panickssery for generously sharing her code implementation, which greatly facilitated this work.

\bibliographystyle{unsrt}  
\bibliography{references}

\begin{thebibliography}{10}

\bibitem{nostalgebraist2020interpreting}
nostalgebraist.
\newblock Interpreting {GPT}: the logit lens.
\newblock \url{https://www.lesswrong.com/posts/AcKRB8wDpdaN6v6ru/interpreting-gpt-the-logit-lens}, August 2020.
\newblock Accessed: 2025-02-22.

\bibitem{zhaoyi2024understanding}
Li~Zhaoyi, Jiang Gangwei, Xie Hong, Song Linqi, Lian Defu, and Wei Ying.
\newblock Understanding and patching compositional reasoning in llms.
\newblock {\em arXiv preprint arXiv:2402.14328}, 2024.

\bibitem{wendler2024llamasworkenglishlatent}
Chris Wendler, Veniamin Veselovsky, Giovanni Monea, and Robert West.
\newblock Do llamas work in english? on the latent language of multilingual transformers, 2024.

\bibitem{ashkan2024mechanistic}
Golgoon Ashkan, Filom Khashayar, and Kannan Arjun, Ravi.
\newblock Mechanistic interpretability of large language models with applications to the financial services industry.
\newblock {\em arXiv preprint arXiv:2407.11215}, 2024.

\bibitem{panickssery2023decoding}
Nina Panickssery.
\newblock Decoding intermediate activations in llama-2-7b.
\newblock LessWrong, 2023.
\newblock Accessed: 2025-02-22.

\bibitem{belrose2023elicitinglatentpredictionstransformers}
Nora Belrose, Zach Furman, Logan Smith, Danny Halawi, Igor Ostrovsky, Lev McKinney, Stella Biderman, and Jacob Steinhardt.
\newblock Eliciting latent predictions from transformers with the tuned lens, 2023.

\bibitem{ghandeharioun2024patchscopesunifyingframeworkinspecting}
Asma Ghandeharioun, Avi Caciularu, Adam Pearce, Lucas Dixon, and Mor Geva.
\newblock Patchscopes: A unifying framework for inspecting hidden representations of language models, 2024.

\bibitem{katz2024backwardlensprojectinglanguage}
Shahar Katz, Yonatan Belinkov, Mor Geva, and Lior Wolf.
\newblock Backward lens: Projecting language model gradients into the vocabulary space, 2024.

\bibitem{wu2024semantichubhypothesislanguage}
Zhaofeng Wu, Xinyan~Velocity Yu, Dani Yogatama, Jiasen Lu, and Yoon Kim.
\newblock The semantic hub hypothesis: Language models share semantic representations across languages and modalities, 2024.

\bibitem{jaturong2025an}
Kongmanee Jaturong.
\newblock An attempt to unraveling token prediction refinement and identifying essential layers of large language models.
\newblock {\em arXiv preprint arXiv:2501.15054v1}, 2025.

\bibitem{bhalla2025unifyinginterpretabilitycontrolevaluation}
Usha Bhalla, Suraj Srinivas, Asma Ghandeharioun, and Himabindu Lakkaraju.
\newblock Towards unifying interpretability and control: Evaluation via intervention, 2025.

\end{thebibliography}

\end{document}